\documentclass[journal]{IEEEtran}

\usepackage{times}
\usepackage{soul}
\usepackage{url}
\usepackage[hidelinks]{hyperref}
\usepackage[utf8]{inputenc}
\usepackage{graphicx}
\usepackage{amsmath}
\usepackage{amsthm}
\usepackage{booktabs}
\usepackage{algorithm}
\usepackage{algorithmic}
\usepackage{amsfonts}
\usepackage{makecell}
\usepackage{multirow}
\usepackage{color}
\usepackage{float}
\usepackage{bm}
\usepackage{array}
\usepackage{amssymb}
\usepackage{amsfonts}
\usepackage{array,color}
\usepackage{graphicx}
\usepackage{xcolor}
\usepackage{subfigure}
\usepackage{indentfirst}
\usepackage{enumerate}
\usepackage{verbatim}

\ifCLASSINFOpdf

\else

\fi

\begin{document}

\title{Deep High-Resolution Representation Learning for Cross-Resolution Person Re-identification}
%
%
% author names and IEEE memberships
% note positions of commas and nonbreaking spaces ( ~ ) LaTeX will not break
% a structure at a ~ so this keeps an author's name from being broken across
% two lines.
% use \thanks{} to gain access to the first footnote area
% a separate \thanks must be used for each paragraph as LaTeX2e's \thanks
% was not built to handle multiple paragraphs
%

\author{Guoqing~Zhang,~\IEEEmembership{Member,~IEEE,}
        Yu~Ge,
        Zhicheng~Dong,
        Hao~Wang,
        Yuhui~Zheng,
        and~Shengyong~Chen~\IEEEmembership{Senior Member,~IEEE}
% <-this % stops %a space
\thanks{G. Zhang is with the School of Computer and Software, Nanjing University of Information Science and Technology, Nanjing, 210044, China, and also with the Engineering Research Center of Digital Forensics, Ministry of Education, Nanjing University of Information Science and Technology (e-mail: xiayang14551@163.com). (Corresponding author: Yuhui Zheng).}% %<-this % stops a space
\thanks{Y. Ge, Z. Dong, H. Wang, and Y. Zheng are with the School of Computer and Software, Nanjing University of Information Science and Technology, China, 210044 (e-mail: gy1328447669@gmail.com, dzc2000919@gmail.com, btnode3@gmail.com, zheng\_yuhui@nuist.edu.cn).}% <-this % %stops a space
\thanks{S. Chen is with the School of Computer Science and Engineering, Tianjin University of Technology, Tianjin 300384, China (e-mail: sy@ieee.org).}
}

% note the % following the last \IEEEmembership and also \thanks -
% these prevent an unwanted space from occurring between the last author name
% and the end of the author line. i.e., if you had this:
%
% \author{....lastname \thanks{...} \thanks{...} }
%                     ^------------^------------^----Do not want these spaces!

% The paper headers
\markboth{Journal of IEEE Transactions on Image Processing}%
{Shell \MakeLowercase{\textit{et al.}}: Deep High-Resolution Representation Learning for Cross Resolution Person Re-identification}

\maketitle

% As a general rule, do not put math, special symbols or citations
% in the abstract or keywords.
\begin{abstract}
Person re-identification (re-ID) tackles the problem of matching person images with the same identity from different cameras. In practical applications, due to the differences in camera performance and distance between cameras and persons of interest, captured person images usually have various resolutions. We name this problem as Cross-Resolution Person Re-identification which brings a great challenge for matching correctly. In this paper, we propose a Deep High-Resolution Pseudo-Siamese Framework (PS-HRNet) to solve the above problem. Specifically, in order to restore the resolution of low-resolution images and make reasonable use of different channel information of feature maps, we introduce and innovate VDSR module with channel attention (CA) mechanism, named as VDSR-CA. Then we reform the HRNet by designing a novel representation head to extract discriminating features, named as HRNet-ReID. In addition, a pseudo-siamese framework is constructed to reduce the difference of feature distributions between low-resolution images and high-resolution images. The experimental results on five cross-resolution person datasets verify the effectiveness of our proposed approach. Compared with the state-of-the-art methods, our proposed PS-HRNet improves 3.4\%, 6.2\%, 2.5\%,1.1\% and 4.2\% at Rank-1 on MLR-Market-1501, MLR-CUHK03, MLR-VIPeR, MLR-DukeMTMC-reID, and CAVIAR datasets, respectively.  Our code is available at \url{https://github.com/zhguoqing}.
\end{abstract}

% Note that keywords are not normally used for peerreview papers.
\begin{IEEEkeywords}
Cross-resolution person re-identification, super-resolution, high-resolution network, pseudo-siamese framework, deep learning.
\end{IEEEkeywords}

% For peer review papers, you can put extra information on the cover
% page as needed:
% \ifCLASSOPTIONpeerreview
% \begin{center} \bfseries EDICS Category: 3-BBND \end{center}
% \fi
%
% For peerreview papers, this IEEEtran command inserts a page break and
% creates the second title. It will be ignored for other modes.
\IEEEpeerreviewmaketitle

\section{Introduction}
% The very first letter is a 2 line initial drop letter followed
% by the rest of the first word in caps.
%
% form to use if the first word consists of a single letter:
% \IEEEPARstart{A}{demo} file is ....
%
% form to use if you need the single drop letter followed by
% normal text (unknown if ever used by the IEEE):
% \IEEEPARstart{A}{}demo file is ....
%
% Some journals put the first two words in caps:
% \IEEEPARstart{T}{his demo} file is ....
%
% Here we have the typical use of a "T" for an initial drop letter
% and "HIS" in caps to complete the first word.
\IEEEPARstart{P}{erson} re-identification (re-ID)  intends to match person images with the same identity across images
captured by various cameras. Re-ID has become the spotlight in the field of machine learning and computer vision owing to its wide practicability in recent years \cite{1,2,3,4,5,6,7,8,9}. Driven by recent advances of deep learning, existing researches of re-ID focus on designing deep feature extraction networks to improve the matching accuracy of re-ID \cite{10,11,12,13}. Although these approaches have achieved satisfactory performance and alleviated the influence of person pose changes, background clutters or part occlusions to a certain extent, these methods are usually on the basis of the prerequisite that the gallery images and the query images possess the same resolution and sufficient fine-grained details.

However, such prerequisite is difficult to guarantee in practical applications. The problem of matching the same person images with different resolutions is named as Cross-Resolution Person Re-identification \cite{13,14,15,16}.

\begin{figure}[t]
\centering
\includegraphics[scale=0.95]{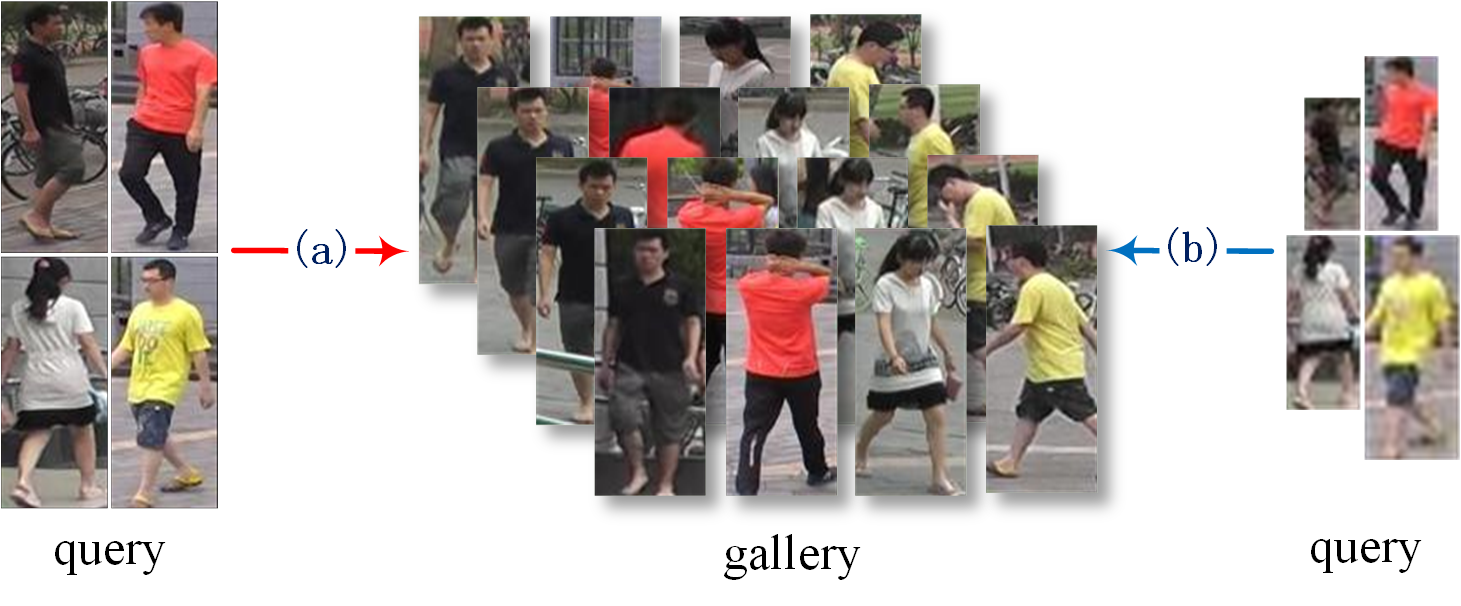}
\caption{Illustration of the difference between (a) traditional person re-ID task in the ideal scenario and (b) cross-resolution person re-ID task. Compared with high-resolution (HR) query images and gallery images, low-resolution (LR) query images contain less fine-grained details, which causes a significant reduction in recognition accuracy and brings great challenges to the work of matching.}
\label{fig:image1}
\end{figure}

Fig. \ref{fig:image1} shows the difference between (a) the person re-ID task in an ideal condition and (b) the cross-resolution person re-ID task. Ideally, query images maintain the same high-resolution (HR) as gallery images. However, due to the differences in camera performance and distance between probes and target pedestrians, captured query images often possess lower resolution than gallery images. The lack of image information makes the traditional re-ID methods incapable of effectively extracting the discriminant features of images for matching, which has become a stumbling block to the development of re-ID task.

In order to settle the above problem, many cross-resolution person re-ID algorithms have been put forward in recent years. In early work, the main idea plans to explore the common feature representation space of HR and LR images by using metric learning or dictionary learning methods, such as \cite{14,13,17,18}. However, the performance of these methods are restricted due to the incapability of recovering the information lost in LR images. Later, some researchers try to introduce super-resolution (SR) technology into cross-resolution person re-ID. SING \cite{16} first applies SRCNN \cite{19} as the resolution recovery module and jointly trains the SRCNN sub-network and the re-ID sub-network. Since then, different SR networks, such as SRGAN \cite{20} and FFSR \cite{21}, are introduced as the resolution recovery module to further optimize the framework. Recently, some new methods represented by INTACT \cite{40} have been proposed, and more novel and effective mechanisms have been applied to raise the detection accuracy to a new level. These methods have achieved significant performance improvement, but it is still far below the practical application standard.

Through detailed comparison and analysis of numerous recent cross-resolution person re-ID methods, we gradually discovered some commonalities contained in them. The most conspicuous is that the existing approaches almost all use convolutional neural networks with the property of down-sampling such as ResNet \cite{53} as feature extraction networks. We believe that this is the most detrimental factor in the existing methods. Using such networks as the feature extraction backbone will inevitably cause further loss of
fine-grained information from low-resolution images. Besides, excessive emphasis on low-resolution image reconstruction seems to have formed a stereotyped thinking pattern. In fact, through experiments, we found that complex super-resolution networks may not perform better than a simple one in cross-resolution person re-ID task under certain circumstances. More energy should be devoted to the study of deep semantic information and feature information extracted from low-resolution images.

In this paper, we propose the PS-HRNet to solve the limitations analyzed above. Firstly, we further improve the super-resolution capability of VDSR \cite{22} in terms of deep semantic information learning by adding the channel attention mechanism, and name the modified SR module as VDSR-CA. Besides, based on the finding that the unique parallel architecture of HRNet \cite{23} is helpful to alleviate the impact of resolution difference, we utilize the HRNet as the feature extraction network. Here we propose the HRNet-ReID to capture multi-resolution features of person images by introducing a novel representation head to HRNet, which can adapt HRNet to the person re-ID problem. In addition, our PS-HRNet adopts a pseudo-siamese framework \cite{65,66} so as to further decrease the distribution difference between LR image features and HR image features. The training strategy of the whole network is divided into two phases. In the first phase, only the HRNet-ReID module in HR branch of pseudo-siamese framework is trained on traditional HR person re-ID datasets. In the second phase, we use joint training strategy to train both the VDSR-CA module and two HRNet-ReID modules simultaneously on  cross-resolution person re-ID datasets.

The outstanding contributions of our work are summarized in the following three points:

%$\bullet$ We propose a modified version of VDSR named as VDSR-CA by combining the channel attention mechanism, and further improve the super-resolution capability while maintaining a low-complexity structure.

$\bullet$ We put forward a feature extraction network named as HRNet-ReID, which combines native HRNet-W32 backbone with a novel representation head designed by us to adapt HRNet to the specific person re-ID mission, overcoming the flaw caused by conventional feature extraction networks in existing methods.

$\bullet$ We construct a pseudo-siamese framework named as PS-HRNet which combines our proposed VDSR-CA and HRNet-ReID to further explore the feature space at a deeper level and successfully reduce the distribution difference between LR image features and HR image features, providing an original solution to the cross-resolution person re-ID problem.

$\bullet$ We have carried out extensive experiments on five cross-resolution person re-ID datasets, and all of them have achieved the highest level in the industry. Compared with the state-of-the-arts, our proposed PS-HRNet improves 3.4\%, 6.2\%, 2.5\%, 1.1\% and 4.2\% at Rank-1 on MLR-Market-1501, MLR-CUHK03, MLR-VIPeR, MLR-DukeMTMC-reID, and CAVIAR datasets, respectively.

\begin{figure*}[t]
\centering
\includegraphics[scale=0.79]{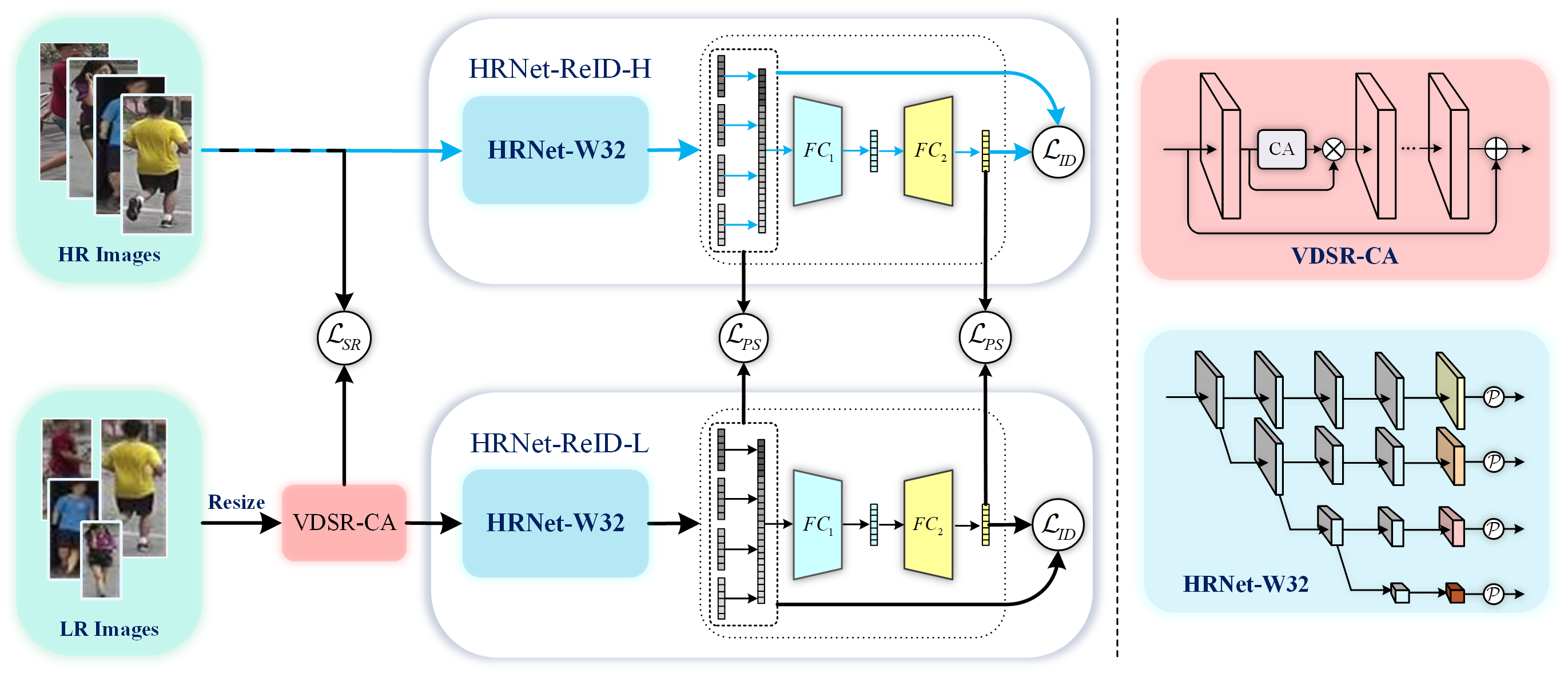}
\caption{The unified architecture of Deep High-Resolution Pseudo-Siamese Framework (PS-HRNet) proposed by us for cross-resolution person re-ID task, which is a pseudo-siamese framework consisting of double HRNet-ReID feature extraction networks and one VDSR-CA module. The VDSR-CA is utilized to restore the resolution of input LR images. The HRNet-ReID is designed to adapt the HRNet to the person re-ID task, and extracts discriminating features from restored images. The right side gives a brief sketch of the two types of modules. $\otimes$ and $\oplus$ denote element-wise product and element-wise add, respectively. As the main structure in PS-HRNet, the pseudo-siamese framework is adopted to close the feature distributions of LR and HR images. Our PS-HRNet is trained alternatively in two phases: \textbf{(1)} Update the single HRNet-ReID-H with the loss ${{\cal L}_{ID}}$ (Eq. (10)); \textbf{(2)} Update the joint multi-task learning loss with the loss ${{\cal L}_{TOTAL}}$ (Eq. (12)). The above two phases are marked with \textcolor[RGB]{0,176,240}{blue} and \textbf{black} arrows, respectively. (best viewed in color).}
\label{fig:image2}
\end{figure*}

\section{RELATED WORK}
In this section, we will roughly introduce the related works concerned with traditional person re-ID and cross-resolution person re-ID, and revisit two utilized core modules and effective pseudo-siamese framework.

\subsection{Person re-ID}
Person re-ID has achieved rapid development in the past decades. A series of methods have been proposed to extract more robust and discriminating feature representations, and overcome the difficulties brought by person pose changes, background clutters or part occlusions. Specifically, to solve pose changes, Liu et al. \cite{26} design a pose-transferable GAN which aims to produce person images with multiple poses for data enhancement. To address background clutter, some methods based on attention mechanism or semantic parsing are proposed. Li et al. \cite{27} apply spatial and channel attention to make network focus on more informative parts. Kalayeh et al. \cite{28} adopt semantic parsing to segment the foreground information and background information to reduce the interference of background. Besides, extensive methods have achieved great progress in occluded re-ID \cite{29}, \cite{30}, unsupervised re-ID \cite{31}, \cite{32}, cross-modality re-ID \cite{33}, \cite{34}, and so on. However, most of existing methods neglect the resolution mismatch problem which is a common situation in practical scenarios.

\subsection{Cross-Resolution Person re-ID}
To solve the resolution inconsistency problem, a few methods have been proposed recently. Previous traditional methods \cite{35}, \cite{36} mainly focus on dictionary learning and metric learning, which achieve limited performance due to the lack of detail features in LR images. Encouraged by the flourishing development of convolutional neural networks (CNN) \cite{37} and super-resolution (SR) technology, some SR-based methods are proposed and greatly improve the matching accuracy. For instance, Jiao et al. \cite{16} make the first attempt to combine the SRCNN and re-ID network into one framework, and propose a jointly training strategy. Besides, some methods adopt GANs to further improve the framework. Specifically, Wang et al. \cite{38} adopt SR-GAN repeatedly to build a cascaded structure. Li et al. \cite{39} restore image resolutions and learn the resolution-invariant representations. Recently, Cheng et al. \cite{40} optimize SR-reID joint framework from the perspective of training strategy and achieve the best performance, which enhances the compatibility between two sub-networks by utilizing the underlying association knowledge between SR and re-ID.  Most existing deep learning methods based on SR attach their importance to reconstruct SR images and make the generated images visually closer to the original HR images. However, these methods ignore the distribution differences between the LR image features and HR image features extracted by the feature extraction network separately.

\subsection{Revisit VDSR and HRNet}
As a high-performance super-resolution method, VDSR \cite{22} applies a deeper network to gain in-depth image information and further ameliorates the structure of SRCNN \cite{19}. Motivated by the prevalence and development of residual network (ResNet) , VDSR adopts the residual connection to overcome the difficulty in convergence of deep networks. Therefore, the capacity of VDSR on image reconstruction surpasses SRCNN strikingly.

HRNet \cite{23} is first proposed to deal with the Human Pose Estimation task, and then surpasses all predecessors in other fields such as key point detection, pose estimation and multi-person pose estimation \cite{41}, \cite{42}. The core structure of HRNet contains four parallel streams, which is logically presented as follow:
\begin{small}
\begin{equation}
\begin{aligned}
\begin{array}{r}
{{\cal N}_{11}} \to {{\cal N}_{21}} \to {{\cal N}_{31}} \to {{\cal N}_{41}} \\
{\rm{     }} \searrow {{\cal N}_{22}} \to {{\cal N}_{32}} \to {{\cal N}_{42}} \\
{\rm{                }} \searrow {{\cal N}_{33}} \to {{\cal N}_{43}} \\
{\rm{                       }} \searrow {{\cal N}_{44}},
\end{array}
\end{aligned}
\end{equation}
\end{small}where ${{\cal N}_{sr}}$ is a sub-stream in the $s$-th phase and $r$ is a resolution index. In order to maintain the high resolution of the same branch while obtaining information from other branches, layers at the junction of different ${\cal N}$ in the same $s$-th can propagate all of its information to every sub-stream ${{\cal N}}$ in the next $s$-th phase. Such unique parallel-cross structure enables the module to acquire and transmit information between networks of different scales. Besides, the structure has the ability to maintain high-resolution feature map for each branch and each phase.

\subsection{Pseudo-Siamese Framework}

The siamese neural network architecture was first proposed to verify the signature of a check at the end of the last century, and achieved satisfactory expectations \cite{24}. It feeds inputs into two identical neural networks that share weights with each other to map the inputs to a brand new feature space and then measure the difference between the inputs under the new feature representation \cite{25}.

Inspired by the success of siamese neural network architecture, the pseudo-siamese framework is proposed for various detection and recognition tasks \cite{66}. Different from the weights sharing mechanism of the siamese neural network architecture, the network contained in the latter does not need to share weights, so it can be composed of two identical or different sub-networks, which makes the pseudo-siamese framework
possess higher degrees of flexibility and a wider range of application scenarios. The proposal of the pseudo-siamese network brings a novel thinking to traditional classification and comparison tasks.

On the basis of the above existing works, in our method, we add the channel attention (CA) mechanism \cite{43} to VDSR, which makes the network perceive the more informative channels in the process of image reconstruction. In addition, we design a brand-new representation head of HRNet to make full utilization of person image features. Finally, we structure the pseudo-siamese framework, composed of two different sub-networks, by which high-resolution and low-resolution pedestrian images received respectively to explore the similarity and feature distribution of cross-resolution images in new feature representation.

\section{PROPOSED METHOD}
In this section, the proposed PS-HRNet is introduced for the cross-resolution person re-ID problem by giving an overview of its unified architecture first, followed by the details of its main components and the pseudo-siamese framework.

\subsection{Framework Overview}
As illustrated in Fig. \ref{fig:image2}, our proposed PS-HRNet adopts a pseudo-siamese framework as the global structure, and contains two main modules, i.e., HRNet-ReID and VDSR-CA. In order to clearly clarify the following formulas, we first define the notations of datasets which will be used in this paper. In the training phase, we define $N$ high-resolution (HR) images with associated labels as ${{\cal D}_h} = \{ x_h^i,y^i\} _{i = 1}^N$, where $x_h^i \in {\mathbb{R} ^{H \times W \times 3}}$. We down-sample each HR image with the down-sample rate $r \in \left\{ {2,3,4} \right\}$ (i.e. the spatial size of a down-sampled image becomes $\frac{H}{r} \times \frac{W}{r}$). The generated corresponding low-resolution (LR) images are denoted as $ {{\cal D}_l} = \{ x_l^i,y^i\} _{i = 1}^N$.

Our proposed PS-HRNet method has two main objectives. Just as most SR-based methods, one objective is to use the super-resolution reconstruction module to restore the missing detail information in LR images and reduce the visual difference between LR and HR images. In our method, the original VDSR \cite{22} is improved by adding a channel attention block and named as VDSR-CA, which is applied to generate a HR version for each LR image. The other is to minimize the discrepancies in feature distribution between LR and HR images and enhance the matching accuracy by constructing a pseudo-siamese framework. For each pair of input images, we utilize HRNet-ReID as the feature extraction network for each branch to learn HR and LR feature representations, and use losses to promote the reduction in distribution differences of these features. These modules will be illustrated meticulously in following subsections.

\subsection{VDSR with Channel Attention}
In most traditional CNN-based super-resolution modules, each channel of the feature map is treated equally in the process of information transmission between every two feature maps. However, the reality is that the image features contained in different channels of the feature map are various, and these diversities contribute to the recovery of high-frequency features in super-resolution task to a different extent. So it is very meaningful to assign different weights to the channels of feature map to embody the difference between the channels. Therefore, we adopt a channel attention (CA) mechanism proposed in RCAN \cite{43} to redistribute the characteristics of different channels of the
feature map to enhance the recovery ability of SR networks. The formulaic representation of the CA mechanism is as follows:
\begin{small}
\begin{equation}
\begin{aligned}
{\hat f_c} = {s_c} \cdot {f_c},
\end{aligned}
\end{equation}
\end{small}where ${f_c}$ and ${s_c}$ denote the feature map and scaling factor in the same $c$-th channel of input image. The complete expression of $s$ is
\begin{small}
\begin{equation}
\begin{aligned}
s = S\left( {{W_U}\delta \left( {{W_D}z} \right)} \right),
\end{aligned}
\end{equation}
\end{small}where ${W_U}$ and ${W_D}$ denote the channel-upscaling layer and channel-downscaling layer with the ratio $r$ as the sampling rate, respectively. $S$ and $\delta $ denote Sigmoid gating \cite{44} and the activation function ReLU \cite{45}, respectively. The channel-wise statistic $z \in \mathbb R{^C}$ is obtained from the feature map and the $c$-th element of $z$ is represented by the following formula:
\begin{small}
\begin{equation}
\begin{aligned}
{z_c} = {H_{GP}}({f_c}) = \frac{1}{{H \times W}}\sum\limits_{i = 1}^H {\sum\limits_{j = 1}^W {{f_c}\left( {i,j} \right)} },
\end{aligned}
\end{equation}
\end{small}where $H_{GP}$ denotes global pooling function, and ${{f_c}\left( {i,j} \right)}$ denotes the pixel value at position $({i,j})$ of $c$-th feature ${f_c}$.

On the basis of RCAN's contribution, we adopt the VDSR as the super-resolution (SR) module which is fused with the channel attention (CA) mechanism to further enhance the performance. We add CA layer between every two layers of the original VDSR network. The fused SR module is named as VDSR-CA. The simplest Manhattan distance is utilized to train the VDSR-CA module, and the loss ${{\cal L}_{SR}}$ is calculated as:
\begin{small}
\begin{equation}
\begin{aligned}
{{\cal L}_{SR}} = \sum\limits_{i = 1}^{P \times K} {||{\mathbb E}{_{x_l^i\sim{D_l},x_h^i\sim{D_h}}}[{\cal G}\left( {x_l^i} \right) - x_h^i]|{|_1}},
\end{aligned}
\end{equation}
\end{small}where ${\cal G}$ denotes VDSR-CA module. $P$ and $K$ denote the number of selected persons and the number of corresponding images of each selected person, respectively.

\subsection{Innovation of High-Resolution Network}

HRNet aims to fully extract features of input person images for retrieving and matching. Existing representation head of HRNet, such as HRNet-W32-C cannot perform satisfactorily in re-ID task. For this reason, we design a new representation head of HRNet which is adapted to re-ID task. We name the improved HRNet as HRNet-ReID.
\begin{figure}[!t]
\centering
\includegraphics[height=5.2cm]{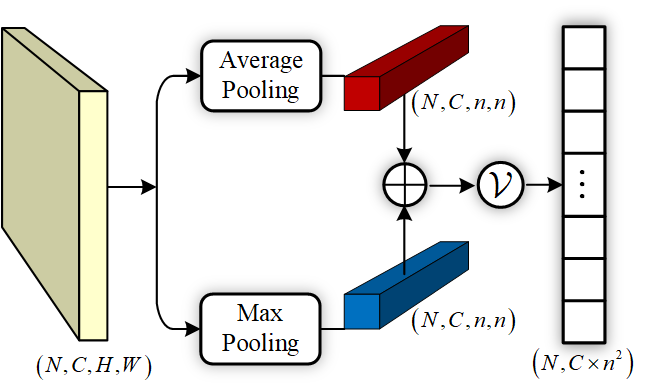}
\caption{The detailed process of the last feature map in each branch of HRNet-ReID. Both Adaptive Average Pooling(AAP) and Adaptive Max Pooling(AMP) are utilized to extract the channel features from the output of each branch. Then the two obtained features are added together, and further reshaped into a feature sequence.}
\label{fig:image3}
\end{figure}

Fig. \ref{fig:image3} illustrates the elaborate processing of the last feature map in each branch. For the reason that feature maps with higher resolutions may contain more pixel space information, we design a multi-resolution feature fusion strategy in HRNet. Adaptive Average Pooling (AAP) and Adaptive Max Pooling (AMP) operations are used to compress and refine the feature map information. For each branch, the two feature maps extracted by pooling layers are added into one feature map. Then the final sequence of each branch is obtained by reshaping the generated feature map. Specifically, the output sequence $Se{q^{(n)}}$ is defined as:
\begin{small}
\begin{equation}
\begin{aligned}
Se{q^{(n)}} &= {\cal P}({f^{(n)}},n;AAP,AMP,{\cal V})
\\
&= {\cal V}(AAP({f^{(n)}},n,n) + {\lambda _n}AMP({f^{(n)}},n,n)),
\end{aligned}
\end{equation}
\end{small}where $\cal P$ denotes the mapping from feature map to feature sequence. ${f^{(n)}}$ denotes the final feature map of HRNet in each branch. $n \in \{ 1,2,3,4\}$ denotes the index of branch in HRNet, and ${\cal V}$ denotes the reshape operation which can transfer a feature map into a feature sequence. In particular, ${f^{(n)}}$ also represents the feature map containing the $n{\rm{ - th}}$ highest resolution in the four branches, and the parameter $n$ also controls the output size of adaptive pooling layers.

Guided by the above designment, we use map ${\cal P}$ to serialize the four feature maps of HRNet and reconstruct the feature representations.

\begin{figure}[t]
\centering
\includegraphics[scale=0.92]{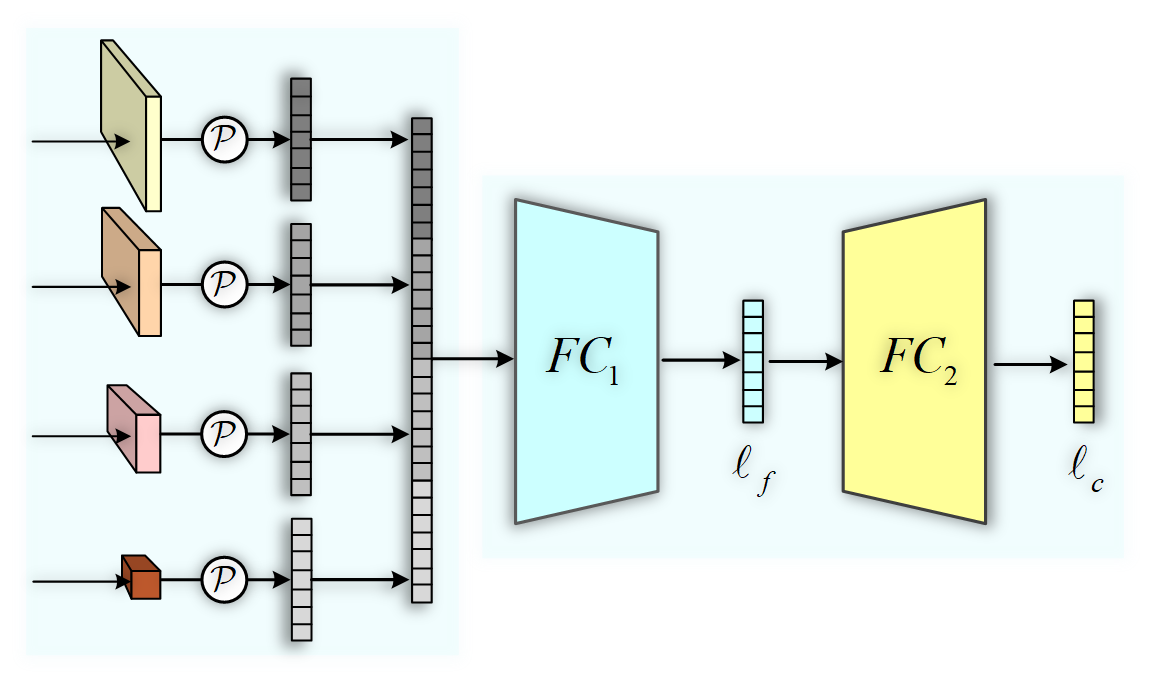}
\caption{Illustration of the proposed representation head in HRNet-ReID. For each branch, the output is first processed by the mapping ${\cal P}$ which is detailed introduced in Fig. \ref{fig:image3}. The extracted feature sequences $Se{q^{(1\sim4)}}$ represented by four short columns of different gray levels are concatenated together and denoted as the longest $Se{q^{(5)}}$. Then we obtain the feature representation ${\ell _f}$ and the classification output ${\ell _c}$ through two full connection layers $F{C_1}$ and $F{C_2}$.}
\label{fig:image4}
\end{figure}

Fig. \ref{fig:image4} shows the architecture of the entire output representation which contains $Se{q^{(1\sim4)}}$ obtained from map ${\cal P}$ and $Se{q^{(5)}}$ concatenated by $Se{q^{(1\sim4)}}$. We exploit a simple classification module to further process $Se{q^{(5)}}$ which consists of two fully connected layers denoted as $F{C_1}$ and $F{C_2}$. The outputs of them are represented as ${\ell _f}$ and ${\ell _c}$ respectively.% In particular, when the input image is HR, we call them ${\ell _{{f_h}}}$ and ${\ell _{{c_h}}}$, and when the input image is LR, we call them ${\ell _{{f_l}}}$ and ${\ell _{{c_l}}}$.

\subsection{HRNet-ReID Training}
In recent years, a unified learning strategy which combines metric learning and representation learning has been widely applied in the person re-ID task and achieved great performance. In the training phase, the extracted $Se{q^{(1\sim5)}}$ in HRNet-ReID participate in metric learning, and the classification output ${\ell _c}$ participates in the representation learning. In the testing phase, we concatenate $Se{q^{(5)}}$ and ${\ell _f}$ as the final feature representation for evaluation.

A representative loss function in metric learning called triplet loss is usually used for fine-grained recognition at the individual level. Here, we take advantage of the batch hard triplet loss ${{\cal L}_{BH}}$, which is calculated as:
\begin{small}
\begin{equation}
\begin{aligned}
\begin{array}{l}
{{\cal L}_{BH}} = \sum\limits_{i = 1}^P {\sum\limits_{a = 1}^K {\sum\limits_{t = 1}^5 {[m + {{\max }_{p = 1...K}}||Seq_{a,i}^{(t)} - Seq_{p,i}^{(t)}|{|_2}} }}
\\
\quad\quad\quad\quad\quad\quad\quad\quad\quad - {\min _{\scriptstyle j = 1...P\hfill\atop
{\scriptstyle n = 1...K\hfill\atop
\scriptstyle j \ne a\hfill}}}||Seq_{a,i}^{(t)} - Seq_{n,j}^{(t)}|{|_2}{]_ + },
\end{array}
\end{aligned}
\end{equation}
\end{small}where $a$ denotes an anchor image, $p$ denotes a positive sample image, $n$ denotes a negative sample image, and $m$ represents a margin parameter to control the differences between intra and inter distances.

Moreover, we adopt the cross entropy label smooth loss that combines a label smoothing mechanism as the classification loss ${{\cal L}_{CE}}$ for representation learning:
\begin{small}
\begin{equation}
\begin{aligned}
{{\cal L}_{CE}} = \sum\limits_{n = 1}^{P \times K} {[ - \sum\limits_{y = 1}^M {\log (p(y))q(y)} ]},
\end{aligned}
\end{equation}
\end{small}where $M$ represents the number of person labels involved in training set, and $p(y)$ denotes the probability that predicted label is $y$. Besides, the definition of $q(y)$ is:
\begin{small}
\begin{equation}
\begin{aligned}
q(y) = \left\{ \begin{array}{l}
1 - \frac{{M - 1}}{M}\varepsilon \quad\quad if\quad y = {y_{truth}}\\\\
\frac{\varepsilon }{M} \quad\quad\quad\quad\quad\quad others
\end{array} \right.
\end{aligned}
\end{equation}
\end{small}where ${y_{truth}}$ is the ground-truth label of the input image and $\varepsilon$ is a parameter of ${{\cal L}_{CE}}$.

Based on the two loss functions mentioned above (Eq. (7) and (8)), for each batch of training set, we compute the HRNet-ReID loss ${{\cal L}_{ID}}$ by:
\begin{small}
\begin{equation}
\begin{aligned}
{{\cal L}_{ID}} = {\lambda _{CE}}{{\cal L}_{CE}} + {\lambda _{BH}}{{\cal L}_{BH}},
\end{aligned}
\end{equation}
\end{small}where ${\lambda _{CE}}$ and ${\lambda _{BH}}$ are parameters to control the importance of ${{\cal L}_{CE}}$ and ${{\cal L}_{BH}}$, respectively.

It is worth emphasizing that, in the whole architecture of our PS-HRNet, HRNet-ReID as a feature extraction network, which plays different roles in different phases. As shown in Fig. \ref{fig:image2}, HRNet-ReID undertakes the task of effectively extracting discriminating features from input HR images in the first phase of the blue arrow representation. In the second phase of the black arrow representation, HRNet-ReID of second phase both participates in multi-task joint learning and forms a pseudo-siamese framework with the first phase HRNet-ReID. For the convenience of representation and distinction, when the resolution of the input images is HR or LR, we define HRNet-ReID as HRNet-ReID-H and HRNet-ReID-L, respectively.

\subsection{Multi-Task Learning Under Pseudo-Siamese Framework}

\begin{table}[!t]
  \centering
     \setlength{\tabcolsep}{1mm}{
    \begin{tabular}{l}
    \toprule
    \textbf{Algorithm 1 PS-HRNet module training} \\
    \midrule
    \textbf{Input:} Training set ${{\cal D}_h} = \{ x_h^i,y^i\} _{i = 1}^N$ and ${{\cal D}_l} = \{ x_l^i,y^i\} _{i = 1}^N$. \\\\
    \textbf{Output:} Joint re-ID module composed of \textbf{VDSR-CA} and \textbf{HRNet-ReID-L}. \\\\
    \textbf{Phase 1 (Preparation)} \\
    \quad Take ${{\cal D}_h}$ as input. \\
    \quad \textbf{for} $i=1$ \textbf{to} $iter_1$ \textbf{do} \\
    \quad \quad Update the single HRNet-ReID-H with the loss ${{\cal L}_{ID}}$ (Eq. (10)). \\
    \quad \textbf{end for} \\\\
    \textbf{Phase 2 (Joint Multi-Task Learning)} \\
    \quad Take ${{\cal D}_h}$ and ${{\cal D}_l}$ as input. \\
    \quad Import the first phase trained HRNet-reID-H module. \\
        \quad \textbf{for}  $i=1$ \textbf{to} $iter_2$ \textbf{do} \\
        \qquad Update the joint multi-task learning loss with the loss ${{\cal L}_{TOTAL}}$\\\qquad (Eq. (12)). \\
    \quad \textbf{end for} \\
    \bottomrule
    \end{tabular}%
    }
  \label{tab:addlabel}%
\end{table}%

The  design  and  application  of  the  pseudo-siamese  frame-work  make  the  training  mode  of  PS-HRNet  different  from most other cross-resolution person re-ID methods. To build a joint multi-task learning structure as the black arrow shown in Fig. 2, we specially design a set of training strategies.

In the first phase, single HRNet-ReID is trained individually with the HR images from ${\cal D}_h$. By minimizing the loss ${\cal L}_{ID}$ and observing the indicators of HRNet-ReID in the testing set, we obtain a high-performance HRNet-ReID module defined as HRNet-ReID-H for subsequent joint learning and construction of the pseudo-siamese framework.

In the second phase, we first concatenate VDSR-CA with the newly defined HRNet-ReID-L, and then construct the pseudo-siamese framework with the obtained HRNet-ReID-H from the first phase and HRNet-ReID-L from the concatenated structure via the Manhattan distance loss ${\cal L_{PS}}$. The pseudo-siamese framework is adopted to measure the similarity of two inputs. Compared with the siamese framework, the pseudo-siamese framework is more suitable for the situation where two inputs have a certain difference. We first define the ordered set ${A} = \left\{ {Seq^{(1)},Seq^{(2)},Seq^{(3)},Seq^{(4)},Seq^{(5)},{\ell _{{c}}}} \right\}$. Then the loss ${\cal L_{PS}}$ is defined as:
\begin{small}
\begin{equation}
\begin{aligned}
{\cal L_{PS}} = \sum\limits_{i = 1}^m {||C_h^{(i)} - C_l^{(i)}} |{|_1},
\end{aligned}
\end{equation}
\end{small}where ordered set ${C} \subseteq {A}$ and ${C} \ne \varnothing$, denotes a combination of elements in ordered set $A$. $C_h$ and $C_l$ denote the set of elements from HRNet-ReID-H and HRNet-ReID-L under the same combination, respectively. $m$ denotes the amount of elements in ordered set $C$.
The process of joint multi-task learning depends on the training set from both ${\cal D}_h$ and ${\cal D}_l$. The VDSR-CA module first reads the LR images from ${\cal D}_l$ for super-resolution reconstruction, and outputs the restored images with the same resolution as the HR images. The restored images are read by HRNet-ReID-H to train itself with the loss ${{\cal L}_{ID}}$ (Eq. (10)), and at the same time, they participate in the calculation of ${{\cal L}_{SR}}$ (Eq. (5)) together with the corresponding HR images from ${\cal D}_h$. Finally, the guidance of HRNet-ReID-H to HRNet-ReID-L is realized by loss ${\cal L_{PS}}$ (Eq. (11)).

To sum up, we combine the ID loss (Eq. (10)), the SR loss (Eq. (5)) and the PS loss (Eq. (11)) into the joint multi-task learning loss ${{\cal L}_{TOTAL}}$, defined as:
\begin{small}
\begin{equation}
\begin{aligned}
{{\cal L}_{TOTAL}} = {{\cal L}_{I{D}}} + {\lambda _{SR}}{{\cal L}_{{\rm{SR}}}}{\rm{ + }}{\lambda _{PS}}{{\cal L}_{PS}},
\end{aligned}
\end{equation}
\end{small}where ${\lambda _{SR}}$ and ${\lambda _{PS}}$ are two parameters to control the importance of ${{\cal L}_{{\text{SR}}}}$ and ${{\cal L}_{PS}}$, respectively. Algorithm 1 gives a summary of the entire training process.

When in the testing phase, we uniformly input the gallery and query images into the trained network composed of VDSR-CA and HRNet-ReID-L. The generated $Se{q^{(5)}}$ and ${\ell _f}$ are concatenated as the ultimate feature representation for matching and evaluation. The entire testing process is executed end-to-end without additional operations.

\section{EXPERIMENTS}
In this part, we first give detailed descriptions of the datasets for evaluation, the experimental settings, and the specific implementation details. Then, a number of experiments are carried out to prove the validity of our proposed method through the comparison with existing methods and ablation studies.

\subsection{Datasets}
Our experiment involves nine datasets, including four high-resolution re-ID datasets for traditional re-ID task: VIPeR \cite{46}, CUHK03 \cite{47}, DukeMTMC-reID \cite{48}, Market-1501 \cite{49}, and four synthetic cross-resolution datasets named as MLR datasets which are constructed from traditional versions: MLR-VIPeR, MLR-CUHK03, MLR-DukeMTMC-reID, MLR-Market1501, as well as one native cross-resolution dataset sampled in real world: CAVIAR \cite{50}. The five cross-resolution datasets involved in our experiments are described as follows:

\subsubsection{MRL-VIPeR}
The MLR-VIPeR includes 632 person-image pairs captured from 2 cameras, a total of 1264 pictures.  Following \cite{16}, we randomly down-sample all the pictures captured by one of the cameras by a down-sampling rate $r \in \{ 2,3,4\}$ and the images collected from another camera remain unchanged.  Here we use the standard 316/316 training/testing identity split.

\subsubsection{MLR-CUHK03}
The MLR-CUHK03 dataset is generated from images taken by 10 (5 pairs) different cameras. It contains 14097 images of 1467 individuals. As \cite{16}, for each pair of cameras, we down-sample the images captured from one camera by randomly selecting a down-sampling rate $r \in \{ 2,3,4\}$, while the resolution of the images collected by the other cameras are unchanged. Here we use the 1,367/100 training/testing identity split.

\subsubsection{MLR-DukeMTMC-reID}
The MLR-DukeMTMC dataset includes 36,411 images of 1,404 identities captured from 8 cameras. Following \cite{39}, we randomly select one camera, and down-sample the images by the same down-sampling rate while the image resolution of other cameras remains unchanged. We use the standard 702/702 training/testing identity split.

\subsubsection{MLR-Market-1501}
The MLR-Market-1501 dataset is composed of 32,668 pictures of 1,501 persons captured by 6 cameras.  Following \cite{39}, we preprocess images of one camera with the same down-sampling rate, while other image resolutions remain unchanged. According to the person ID label, the dataset is separated into a training set containing 751 pedestrians and a testing set containing 750 pedestrians.

\subsubsection{CAVIAR}
The CAVIAR is a dataset collected in real world. It consists of 1,220 pictures of 72 persons collected from 2 cameras. Following \cite{16}, 22 persons are discard by us with only HR images and we randomly split the dataset into two halves based on 25 identities labels for training and testing, respectively.

\renewcommand\arraystretch{1.3}
\begin{table*}[!t]
  \centering
  \caption{Experimental results of cross-resolution person re-ID (\%). The bold and underlined numbers indicate top two results, respectively.}
    \setlength{\tabcolsep}{1.3mm}{
    \begin{tabular}{l|ccc|ccc|ccc|ccc|ccc}
    \toprule
    \multicolumn{1}{c|}{\multirow{2}[4]{*}{Module}} & \multicolumn{3}{c|}{MLR-Market-1501} & \multicolumn{3}{c|}{MLR-CUHK03} & \multicolumn{3}{c|}{MLR-VIPeR} & \multicolumn{3}{c|}{MLR-DukeMTMC-reID} & \multicolumn{3}{c}{CAVIAR} \\
\cmidrule{2-16}          & Rank1 & Rank5 & Rank10 & Rank1 & Rank5 & Rank10 & Rank1 & Rank5 & Rank10 & Rank1 & Rank5 & Rank10 & Rank1 & Rank5 & Rank10 \\
    \midrule
    PCB \cite{51} & 76.9  & 88.9  & 92.4  & 75.3  & 92.7  & 98.1  & 42.6  & 65.8  & 75.9  & 66.4  & 82.5  & 87.1  & 35.9  & 72.1  & 88.6 \\
    DenseNet-121 \cite{52} & 60    & 78.8  & -     & 70.8  & 91.3  & -     & 31.4  & 63.1  & -     & -     & -     & -     & 31.1  & 65.5  & - \\
    ResNet-50 \cite{53} & 57    & 78.7  & -     & 67.4  & 91.7  & -     & 29.9  & 62.2  & -     & -     & -     & -     & 29.6  & 64    & - \\
    SE-ResNet-50 \cite{54} & 58.2  & 78.6  & -     & 70.8  & 92.3  & -     & 33.5  & 63.6  & -     & -     & -     & -     & 30.8  & 65.1  & - \\
    SPreID \cite{28} & 77.4  & 89    & 93.9  & 76.5  & 92.5  & 98.3  & 42.4  & 65.8  & 75.1  & 68.4  & 84.5  & 89.1  & 36.2  & 71.9  & 88.7 \\
    Part Aligned \cite{55} & 75.6  & 88.5  & 92.2  & 73.4  & 92.1  & 97.5  & 40.2  & 62.3  & 73.1  & 67.5  & 83.1  & 87.2  & 35.7  & 71.4  & 87.9 \\
    CamStyle \cite{56} & 74.5  & 88.6  & 93    & 69.1  & 89.6  & 93.9  & 34.4  & 56.8  & 66.6  & 64    & 78.1  & 84.4  & 32.1  & 72.3  & 85.9 \\
    PyrNet \cite{57} & 83.8  & 93.3  & 95.6  & 83.9  & 97.1  & 98.5  & -     & -     & -     & 79.6  & 88.1  & 91.2  & 43.6  & 79.2  & 90.4 \\
    FD-GAN \cite{58} & 79.6  & 91.6  & 93.5  & 73.4  & 93.8  & 97.9  & 39.1  & 62.1  & 72.5  & 67.5  & 82    & 85.3  & 33.5  & 71.4  & 86.5 \\
    \midrule
    JUDEA \cite{14} & -     & -     & -     & 26.2  & 58    & 73.4  & 26    & 55.1  & 69.2  & -     & -     & -     & 22    & 60.1  & 80.8 \\
    SDF \cite{59} & -     & -     & -     & 22.2  & 48    & 64    & 9.3   & 38.1  & 52.4  & -     & -     & -     & 14.3  & 37.5  & 62.5 \\
    SLD$^2$L \cite{13} & -     & -     & -     & -     & -     & -     & 20.3  & 44    & 62    & -     & -     & -     & 18.4  & 44.8  & 61.2 \\
    SING \cite{16} & 74.4  & 87.8  & 91.6  & 67.7  & 90.7  & 94.7  & 33.5  & 57    & 66.5  & 65.2  & 80.1  & 84.8  & 33.5  & 72.7  & 89 \\
    CSR-GAN \cite{38} & 76.4  & 88.5  & 91.9  & 71.3  & 92.1  & 97.4  & 37.2  & 62.3  & 71.6  & 67.6  & 81.4  & 85.1  & 34.7  & 72.5  & 87.4 \\
    FFSR \cite{60} & 59.2  & 80.1  & -     & 70.5  & 92.3  & -     & 40.3  & 65.3  & -     & -     & -     & -     & 31.1  & 68.7  & - \\
    RIFE \cite{60} & 62.6  & 82.4  & -     & 69.7  & 91.5  & -     & 33.9  & 63.6  & -     &       &       &       & 35.7  & 74.9  & - \\
    FFSR+RIFE \cite{60} & 66.9  & 84.7  &       & 73.3  & 92.6  & -     & 41.6  & 64.9  & -     &       &       &       & 36.4  & 72    &  \\
    RAIN \cite{15} & -     & -     & -     & 78.9  & 97.3  & 98.7  & 42.5  & 68.3  & 79.6  & -     & -     & -     & 42    & 77.3  & 89.6 \\
    CAD-Net \cite{39} & 83.7  & 92.7  & 95.8  & 82.1  & 97.4  & 98.8  & 43.1  & 68.2  & 77.5  & 75.6  & 86.7  & 89.6  & 42.8  & 76.2  & 91.5 \\
    CAD-Net++ \cite{61} & 84.1  & 93    & 96.2  & 83.4  & 98.1  & 99.1  & 43.4  & 68.7  & 78.2  & 77.2  & 88.1  & 90.4  & 43.1  & 76.5  & 92.3 \\
    \midrule
    INTACT \cite{40} & \underline{88.1}  & \underline{95}    & \underline{96.9}  & 86.4  & 97.4  & 98.5  & \underline{46.2}  & \underline{73.1}  & \underline{81.6}  & 81.2  & 90.1  & 92.8  & 44    & 81.8  & 93.9 \\
    PRI \cite{62} & 84.9  & 93.5  & 96.1  & 85.2  & 97.5  & 98.8  & -     & -     & -     & 78.3  & 87.5  & 91.4  & 43.2  & 78.5  & 91.9 \\
    PCB+PRI \cite{62} & \underline{88.1}  & 94.2  & 96.5  & 86.2  & \underline{97.9}  & \underline{99.1}  & -     & -     & -     & 81.6  & 89.6  & \underline{92.4}  & 44.3  & 83.7  & \underline{94.8} \\
    PyrNet+PRI \cite{62} & 86.9  & 93.8  & 96.4  & \underline{86.5}  & 97.7  & \underline{99.1}  & -     & -     & -     & \underline{82.1}  & \textbf{91.1} & 92.8  & \underline{45.2}  & \underline{84.1}  & 94.6 \\
    \midrule
    \textbf{Ours}  & \textbf{91.5} & \textbf{96.7} & \textbf{97.9} & \textbf{92.6} & \textbf{98.3} & \textbf{99.4} & \textbf{48.7} & \textbf{73.4} & \textbf{81.7} & \textbf{82.3} & \underline{90.5}  & \textbf{92.8} & \textbf{48.2} & \textbf{84.5} & \textbf{96.3} \\
    \bottomrule
    \end{tabular}%
    }
  \label{tab:table1}%
\end{table*}%

\subsection{Experimental Settings}
In the training phase, both the four traditional re-ID datasets and the corresponding cross-resolution datasets are applied to train the module. The training process is divided into two phases: In the first phase, we only train the HRNet-ReID-H on the traditional high-resolution re-ID datasets, and obtain a high-performance module; In the second phase, we combine the VDSR-CA module with the HRNet-ReID-L guided by HRNet-ReID-H and jointly train them on both traditional datasets and cross-resolution datasets.

In the testing phase, we evaluate the performance of our proposed PS-HRNet on five cross-resolution datasets, where the query sets contain LR images and the gallery sets contain HR images. In particular, since CAVIAR is a genuine cross-resolution dataset, we follow the experimental setting in \cite{16} for training and testing. We adopt the standard single-shot person re-ID settings, and apply the average cumulative match characteristic (CMC) to quantify the performance and report the results of ranks 1, 5 and 10.

\subsection{Implementation details}
In the VDSR-CA module, we retain the entire structure of the original VDSR network, and embed the Channel Attention (CA) mechanism proposed in the RCAN \cite{43}. The internal parameter $r$ of CA mechanism is set to 4.

With respect to HRNet-ReID module, we select the HRNet-W32-C pretrained with the ImageNet dataset as the backbone for feature extraction. Here we redesign the classifier to adapt to the re-ID task as shown in Fig. \ref{fig:image4}. The lengths of $Se{q^{(1\sim5)}}$ are set to 2048, 1024, 2048, 1024 and 6144, respectively. Besides, the dimension of ${\ell _f}$ is set to 512, and the dimension of ${\ell _c}$ is equal to the number of target categories.

Before training, all images are resized to $256\times128\times3$. A mini-batch contains 24 pairs of images of $P = 4$ persons, and each person has $K = 6$ pairs of HR and LR images. We choose SGD to optimize our module with weight decay $5\times {10^{{\rm{ - }}4}}$. The learning rates for training HRNet-ReID and VDSR-CA are set to $8.5\times {10^{{\rm{ - 3}}}}$ and $8.5\times {10^{{\rm{ - 4}}}}$, respectively, which are decreased by 0.1 every 30 epochs. Our module is trained for 70 epochs in total. The hyper-parameters $m$, ${\lambda _{CE}}$, ${\lambda _{BH}}$, ${\lambda _{SR}}$ and ${\lambda _{PS}}$ are set to 0.1, 1.15, 0.2, 0.5 and 0.5, respectively. In ${{\cal L}_{PS}}$, we select $\left\{ {Se{q^{(1)}},Se{q^{(4)}},Se{q^{(5)}},{\ell _c}} \right\}$ as a combination to participate in the operation. Some data augmentation tricks are utilized, such as random flipping, padding and random cropping. We perform our experiments with PyTorch of version 1.6 on single 11GB NVIDIA RTX 2080Ti GPU.

\subsection{Comparisons to State-of-the-Art Methods}
We compare our PS-HRNet with a series of far-ranging state-of-the-art person re-ID methods, which can be roughly separated into two main categories. (1) Conventional methods designed for traditional person re-ID task: PCB \cite{51}, DenseNet-121 \cite{52}, ResNet-50 \cite{53}, SE-ResNet-50 \cite{54}, SPreID \cite{28}, Part Aligned \cite{55}, CamStyle \cite{56} and FD-GAN \cite{58}; (2) Pointed methods designed for cross-resolution person re-ID task: JUDEA \cite{14}, SDF \cite{59}, SLD$^2$L \cite{13}, SING \cite{16}, CSR-GAN \cite{38}, FFSR \cite{60}, RIFE \cite{60}, FFSR+RIFE \cite{60}, RAIN \cite{15}, CAD-Net \cite{39}, CAD-Net++ \cite{61}, PRI \cite{62}, PCB+PRI \cite{62} and PyrNet+PRI \cite{62}. These methods in the comparison almost cover all the current methods in the cross-resolution re-ID field.

The comparison results of the above approaches on five datasets are listed in Table \ref{tab:table1}. We can evidently observe that:

$\bullet$ Our PS-HRNet obtains state-of-the-art performance on all five datasets, and its Rank-1 outperforms the best competitor by 6.1\% on the MLR-CUHK03 dataset.

$\bullet$ Compared with the conventional person re-ID methods, our method outperforms their best result by 11.9\% on the MLR-Market-1501 dataset, which indicates that the information hided in LR images cannot be extracted and utilized effectively by conventional methods when processing cross-resolution person images. Besides, it also demonstrates that the super-resolution reconstruction module with channel attention mechanism plays a significant role in the cross-resolution re-ID task.

$\bullet$ Compared with the pointed methods designed for solution of cross-resolution person re-ID problem, PS-HRNet outperforms all existing methods, which reflects the importance of the tailor-made high-resolution feature extraction network and pseudo-siamese framework.

\begin{table}[!t]
  \centering
  \caption{Evaluating ours loss components on MLR-Market1501. ID: identity classification loss (Eq. (10)), SR: super-resolution loss (Eq. (5)), PS: our pseudo-Siamese framework loss (Eq. (11)).}
      \setlength{\tabcolsep}{6.4mm}{
       \begin{tabular}{c|ccc}
    \toprule
    Supervision & Rank1 & Rank5 & Rank10 \\
    \midrule
    ID    & 84.7  & 92    & 94.6 \\
    SR+ID & 87.6  & 95.1  & 97 \\
    \midrule
    SR+ID+PS & 91.5  & 96.7  & 97.9 \\
    \bottomrule
    \end{tabular}%
    }
  \label{tab:table2}%
\end{table}%

\subsection{Ablation Study}
\subsubsection{Analysis of Loss Functions}
Our PS-HRNet jointly trains image SR module, feature extraction network and pseudo-siamese framework with several loss functions (cf. Eq. (12) etc.). We use the same research strategy as described in INTACT \cite{40} to study the effectiveness of different losses in PS-HRNet on the MLR-Market-1501 dataset. Table \ref{tab:table2} reports the ablation results, which reflect that:

$\bullet$ When only ID loss is included, compared with Table \ref{tab:table1}, it can be obviously found that the performance of our method surpasses almost all existing methods except INTACT \cite{40} and PCB+PRI \cite{62} on the Rank-1, and even reaches the same performance level of single PRI \cite{62}. The results reflect the great feature extraction ability of HRNet on cross-resolution person images which should be owed to its unique high-resolution parallel structure. This also demonstrates the necessity and rationality of applying HRNet to process low-resolution images.

$\bullet$ With the addition of SR loss, the performance is further improved by 2.9\%, which can prove that the image reconstruction function provided by VDSR-CA module has a positive effect on cross-resolution person re-ID.

$\bullet$ The addition of pseudo-siamese framework loss ultimately increases the Rank-1 by 3.9\%, which verifies the positive effect of the pseudo-siamese framework. Furthermore, it proves the necessity of reducing the discrepancies of feature space between HR and LR images, which has been overlooked all along.

\begin{table}[!t]
  \centering
  \caption{Performance comparison test of super-resolution reconstruction and cross-resolution person reID on the MLR-CUHK03 test set.}
  \setlength{\tabcolsep}{6mm}{
    \begin{tabular}{c|cc|c}
    \toprule
    Module & SSIM  & PSNR  & Rank1 \\
    \midrule
    CycleGAN \cite{63} & 0.55  & 14.1  & 62.1 \\
    SING  \cite{16} & 0.65  & 18.1  & 67.7 \\
    CSR-GAN \cite{38} & 0.76  & 21.5  & 71.3 \\
    CAD-Net \cite{39} & 0.73  & 20.2  & 82.1 \\
    \midrule
    Solo SR & 0.93  & 27.9  & 0.04 \\
    SR+ID & 0.91  & 26.1  & 88.9 \\
    SR+ID+PS & 0.91  & 26.1  & 92.6 \\
    \bottomrule
    \end{tabular}%
    }
  \label{tab:table3}%
\end{table}%

$\bullet$ Following INTACT \cite{40}, experiments on loss are conducted to explore the influence of different loss combinations on image restoration quality and whether it will affect the restored images after using the pseudo-siamese framework. We use PSNR and SSIM which are two quantitative indicators that reflect the quality of image restoration to evaluation. We compare our method with CycleGAN \cite{63}, SING \cite{16}, CSR-GAN \cite{38} and CAD-Net \cite{39}. According to the comparative results in Table \ref{tab:table3}, the solo SR module achieves the
best performance in image super-resolution reconstruction which also confirms the effectiveness of VDSR-CA. After using SR+ID, the PSNR and SSIM are slightly reduced, but the recognition accuracy is greatly improved. Furthermore, after adding the pseudo-siamese framework, the value of PSNR and SSIM have no changed, which indicates that the pseudo-siamese framework does not affect the quality of the restored images on visual level.
\begin{figure*}[!t]
\centering
\includegraphics[width=18cm]{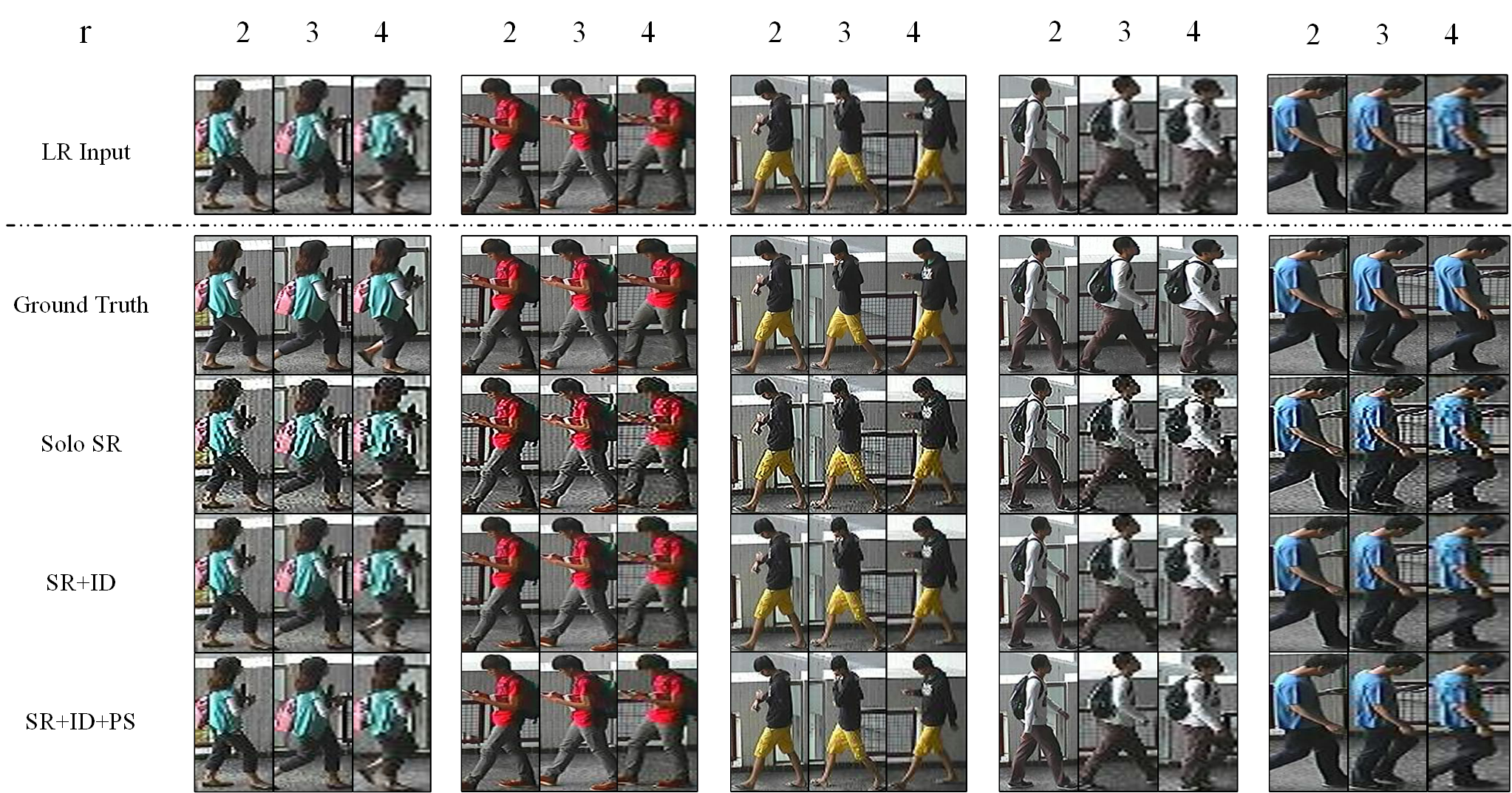}
\caption{Visual comparisons of restored HR test images from the MLR-CUHK03 dataset. Given input images of 3 different low-resolutions $r \in{2,3,4}$, our PS-HRNet outputs the corresponding restored HR images of solo SR module, SR+ID module and SR+ID+PS module.}
\label{fig:image5}
\end{figure*}

$\bullet$ Following INTACT \cite{40}, Fig. \ref{fig:image5} shows some person images restored by our method. These images are selected from the testing set of MLR-CUHK03. Although our PS-HRNet is superior to the existing baseline method in recognition accuracy, there is still a gap between the recovery quality of low-resolution images and the ground truth on visual level. Such outcome does make perfect sense. As we mentioned earlier, the focus is not on the restoration of image quality but on the examination of deep-level feature and semantic information. Blindly using visual sensory experience and related indicators as a standard for image restoration is of little significance because the way the neural network observes the image is completely different from the human visual mechanism. This also gives a reasonable interpretation on the variation of PSNR and SSIM in Table \ref{tab:table3}. Therefore, we do not recommend subsequent studies on cross-resolution person re-identification to conduct excessive super-resolution reconstruction experiments at the aspect of human vision.

\begin{table}[!t]
  \centering
  \caption{Recognition accuracy (\%) of different SR modules on MLR-Market1501.}
        \setlength{\tabcolsep}{4.6mm}{
    \begin{tabular}{c|ccc}
    \toprule
    Module & Rank1 & Rank5 & Rank10 \\
    \midrule
    Bicubic +Resnet50 & 64.7  & 82.1  & 89.1 \\
    SRCNN \cite{19} +Resnet50 & 62.1  & 80.9  & 87.6 \\
    VDSR \cite{22} +Resnet50 & 68.2  & 86.5  & 90.7 \\
    RCAN \cite{43} +Resnet50 & 73.1  & 89.6  & 92.4 \\
    \midrule
    VDSR-CA+Resnet50 & 72.3  & 88.2  & 92 \\
    \bottomrule
    \end{tabular}%
    }
  \label{tab:table4}%
\end{table}%

\subsubsection{Analysis of VDSR-CA}

In order to further explore the effectiveness of the VDSR-CA, we adopt ResNet-50 \cite{53} as a unified feature extraction network on the MLR-Market1501 dataset, and test the effect of using cubic interpolation, SRCNN \cite{19}, VDSR-CA and RCAN \cite{43} as super-resolution modules for joint learning.

The recognition accuracy of different SR modules on MLR-Market1501 are presented in Table \ref{tab:table4}, which reflects that the performance of our VDSR-CA module goes far beyond the classic SRCNN network and is better than the original VDSR network. Although the Rank indicators are slightly lower than the result of joint learning with RCAN, VDSR-CA has a lighter architecture than RCAN which means more practical in real-world applications. Considering the above factors, our VDSR-CA is the best choice for super-resolution modules.

\begin{table}[!t]
  \centering
  \caption{Recognition accuracy (\%) of different pooling method on Market1501 and DukeMTMC-reID.}
  \setlength{\tabcolsep}{0.9mm}{
    \begin{tabular}{c|cccc|cccc}
    \toprule
    \multirow{2}[2]{*}{Method} & \multicolumn{4}{c|}{Market-1501} & \multicolumn{4}{c}{DukeMTMC} \\
          & Rank1 & Rank5 & Rank10 & mAP   & Rank1 & Rank5 & Rank10 & mAP \\
    \midrule
    AAP   & 94    & 97.4  & 98.1  & 83.8  & 86.4  & 93.2  & 95.2  & 73.6 \\
    AMP   & 93.6  & 97.7  & 98.7  & 83.4  & 86.6  & 93.7  & 95.4  & 74 \\
    \midrule
    AAP+AMP & 94.4  & 98.2  & 98.6  & 84.1  & 87    & 94.5  & 96    & 74.3 \\
    \bottomrule
    \end{tabular}%
    }
  \label{tab:table7}%
\end{table}%

\begin{table}[!t]
  \centering
  \caption{Recognition accuracy (\%) of different feature extraction networks on Market1501 and DukeMTMC-reID.}
  \setlength{\tabcolsep}{1.2mm}{
    \begin{tabular}{c|c|cc|cc}
    \toprule
    \multirow{2}[2]{*}{Module} & \multirow{2}[2]{*}{Backbone} & \multicolumn{2}{c|}{Market-1501} & \multicolumn{2}{c}{DukeMTMC-reID} \\
          &       & Rank1 & mAP   & Rank1 & mAP \\
    \midrule
    DPFL \cite{21} & ResNet-50 & 88.9  & 73.1  & 79.2  & 60.6 \\
    PCB \cite{51} & ResNet-50 & 92.3  & 77.4  & 81.7  & 66.1 \\
    FD-GAN \cite{58} & ResNet-50 & 90.5  & 77.7  & 80    & 64.5 \\
    CamStyle \cite{56} & ResNet-50 & 89.2  & 71.6  & 78.6  & 57.6 \\
    DenseNet-121 \cite{52} & DenseNet & 90.2  & 74    & 67.4  & 46.2 \\
    HRNet-W32-C \cite{64} & HRNetV2-W32 & 87.8  & 72.8  & 78.6  & 60.3 \\
    \midrule
    HRNet-ReID  & HRNetV2-W32 & 94.4  & 84.1  & 87    & 74.3 \\
    \bottomrule
    \end{tabular}%
    }
  \label{tab:table5}%
\end{table}%

\subsubsection{Analysis of HRNet-ReID}

The function of HRNet-ReID aims to fully extract features of input person images for retrieving and matching. In order to make a large number of high-dimensional feature map data extracted by the previous HRNet-W32 can be effectively processed by the classification layer at the end of HRNet-ReID, we introduce adaptive average pooling and adaptive max pooling as important means of feature compression.

Table \ref{tab:table7} demonstrates the effect of different pooling options on the recognition accuracy. Obviously, under the same experimental conditions, only the simultaneous application of AAP and AMP can obtain higher detection accuracy. Theoretically, maximum pooling and average pooling play specific roles in extracting feature texture information and global background information respectively. Therefore, the combination of average pooling and adaptive max pooling is valid.

For further investigating the performance of HRNet-ReID as the backbone in the person re-ID task, we conduct experiments on the following conventional person re-ID network combined with ResNet-50, PCB and DenseNet as the comparisons with HRNet-ReID on the Market-1501 and DukeMTMC-reID datasets: DPFL \cite{21}, PCB \cite{51}, FD-GAN \cite{58}, CamStyle \cite{56}, DenseNet121 \cite{52}. In addition, in order to
validate the effectiveness of our designed HRNet representation head, we use the original version of HRNet-W32-C \cite{64} as reference. The representation head of HRNet-W32-C is proposed to solve the problem of image classification on the ImageNet dataset and achieves good performance.

As shown in Table \ref{tab:table5}, the experimental results reflect that our HRNet-ReID significantly outperforms the original HRNet-W32-C and other methods on both datasets which confirms the effectiveness of our modified representation head.

\begin{table}[!t]
  \centering
  \caption{Recognition accuracy (\%) of different sequence combinations on MLR-Market1501.}
  \setlength{\tabcolsep}{1.2mm}{
    \begin{tabular}{c|ccc}
    \toprule
    Sequences & Rank1 & Rank5 & Rank10 \\
    \midrule
    $\left\{ {Se{q^{(5)}},{\ell _c}} \right\}$     & 90.5  & 96.6  & 97.9 \\
    $\left\{ {Se{q^{(1)}},Se{q^{(2)}},Se{q^{(3)}},Se{q^{(4)}},Se{q^{(5)}},{\ell _c}} \right\}$     & 90.9  & 96.1  & 97.5 \\
    \midrule
    $\left\{ {Se{q^{(1)}},Se{q^{(4)}},Se{q^{(5)}},{\ell _c}} \right\}$     & 91.5  & 96.7  & 97.9 \\
    \bottomrule
    \end{tabular}%
    }
  \label{tab:table6}%
\end{table}%

\subsubsection{Analysis of pseudo-siamese framework}
For the purpose of exploring the impact of different training strategies under the pseudo-siamese framework, we select multiple sets of $Se{q^{(n)}}$ and ${\ell _c}$ to participate in the training of pseudo-siamese framework loss and perform testing on the MLR-Market-1501 dataset. According to the combinatorial mathematics, there are 84 combinations for the calculation of Eq. (11) in theory. Here we select three typical combinations for evaluation.

The experimental results are listed in Table \ref{tab:table6}. We can clearly observe that different selections of sequences bring different performances. Here we adopt the combination of $\left\{ {Se{q^{(1)}},Se{q^{(4)}},Se{q^{(5)}},{\ell _c}} \right\}$ in our training strategy. Limited by time, there may be a better combination that further improves the performance of our PS-HRNet.

\section{Conclusion}
In this article, we design a novel approach named Deep High-Resolution Pseudo-Siamese Framework (PS-HRNet) to significantly alleviate the the resolution mismatch problem and improve recognition accuracy in cross-resolution person re-ID task.
Our framework utilizes VDSR-CA as the super-resolution module, HRNet-ReID as the feature extraction network.
The former integrates channel attention mechanism into VDSR, which can reasonably utilize the valuable high frequency components contained in different channels of feature map and restore the missing discriminative information in LR images effectively.
The latter possesses a novel representation head designed by us which can effectively extract fine-grained details from cross-resolution person images.
What's more, the pseudo-siamese framework is adopted and plays a significant component in reducing the distribution difference in feature information between LR and HR images. With extensive experiments, the results confirm that our PS-HRNet can extract discriminating and robust feature representations from cross-resolution images, and achieves the state-of-the-art performance in existing five benchmarks.

% if have a single appendix:
%\appendix[Proof of the Zonklar Equations]
% or
%\appendix  % for no appendix heading
% do not use \section anymore after \appendix, only \section*
% is possibly needed

% use appendices with more than one appendix
% then use \section to start each appendix
% you must declare a \section before using any
% \subsection or using \label (\appendices by itself
% starts a section numbered zero.)
%

%\appendices
%\section{Proof of the First Zonklar Equation}
%Appendix one text goes here.

%\section{}
%Appendix two text goes here.

% use section* for acknowledgment
%\section*{Acknowledgment}

%he authors would like to thank...

% Can use something like this to put references on a page
% by themselves when using endfloat and the captionsoff option.
\ifCLASSOPTIONcaptionsoff
  \newpage
\fi

% trigger a \newpage just before the given reference
% number - used to balance the columns on the last page
% adjust value as needed - may need to be readjusted if
% the document is modified later
%\IEEEtriggeratref{8}
% The "triggered" command can be changed if desired:
%\IEEEtriggercmd{\enlargethispage{-5in}}

% references section

\end{document}